\documentclass[letterpaper]{article} 
\usepackage{aaai2026}  
\usepackage{times}  
\usepackage{helvet}  
\usepackage{courier}  
\usepackage[hyphens]{url}  
\usepackage{graphicx} 
\usepackage{natbib}  
\usepackage{caption} 
\usepackage{algorithm}
\usepackage{algorithmicx}
\usepackage{multirow}
\usepackage{booktabs}
\usepackage{amssymb}
\usepackage{arydshln}
\usepackage{amsmath}
\usepackage{algpseudocode}

\usepackage{newfloat}
\usepackage{listings}
\DeclareCaptionStyle{ruled}{labelfont=normalfont,labelsep=colon,strut=off} 
\floatstyle{ruled}
\newfloat{listing}{tb}{lst}{}
\floatname{listing}{Listing}
\nocopyright

\title{UGD-IML: A Unified Generative Diffusion-based Framework for Constrained and Unconstrained Image Manipulation Localization}
\author{
	Yachun Mi\textsuperscript{\rm 1}, Xingyang He\textsuperscript{\rm 1}, Shixin Sun\textsuperscript{\rm 1}, Yu Li\textsuperscript{\rm 1}, Yanting Li\textsuperscript{\rm 1}, Zhixuan Li\textsuperscript{\rm 2}, \\ Jian Jin\textsuperscript{\rm 2}, Chen Hui\textsuperscript{\rm 1,3}, Shaohui Liu\textsuperscript{\rm 1*} 
}
\affiliations{
	\textsuperscript{\rm 1}Harbin Institute of Technology\\
	\textsuperscript{\rm 2}Nanyang Technological University\\
	\textsuperscript{\rm 3}Nanjing University of Information Science and Technology\\
	
}

\usepackage{bibentry}

\begin{document}

\maketitle

\begin{abstract}
In the digital age, advanced image editing tools pose a serious threat to the integrity of visual content, making image forgery detection and localization a key research focus. Most existing Image Manipulation Localization (IML) methods rely on discriminative learning and require large, high-quality annotated datasets. However, current datasets lack sufficient scale and diversity, limiting model performance in real-world scenarios. To overcome this, recent studies have explored Constrained IML (CIML), which generates pixel-level annotations through algorithmic supervision. However, existing CIML approaches often depend on complex multi-stage pipelines, making the annotation process inefficient.
In this work, we propose a novel generative framework based on diffusion models, named UGD-IML, which for the first time unifies both IML and CIML tasks within a single framework.
By learning the underlying data distribution, generative diffusion models inherently reduce the reliance on large-scale labeled datasets, allowing our approach to perform effectively even under limited data conditions.
In addition, by leveraging a class embedding mechanism and a parameter-sharing design, our model seamlessly switches between IML and CIML modes without extra components or training overhead.
Furthermore, the end-to-end design enables our model to avoid cumbersome steps in the data annotation process.
Extensive experimental results on multiple datasets demonstrate that UGD-IML outperforms the SOTA methods by an average of 9.66\% and 4.36\% in terms of F1 metrics for IML and CIML tasks, respectively.
Moreover, the proposed method also excels in uncertainty estimation, visualization and robustness.


\end{abstract}

\section{Introduction}
The rapid development of technology has given rise to the widespread use of powerful image editing tools such as Photoshop \cite{chen2022self} and FakeApp \cite{zhou2021face}.
Breakthroughs in generative artificial intelligence have further revolutionized image processing, enabling modern AI algorithms to conveniently achieve high-precision image forging \cite{avrahami2022blended,jia2023autosplice,rombach2022high,tailanian2024diffusion}. The popularity of these technologies has dramatically reduced the cost of producing and distributing counterfeit images, posing a serious threat to cybersecurity, information integrity, and personal and organizational security.
Therefore, the development of effective methods for detecting and accurately localizing image forgeries has become a critical and urgent research challenge.

\begin{figure}[t]
	\centering
	\includegraphics[width=7cm]{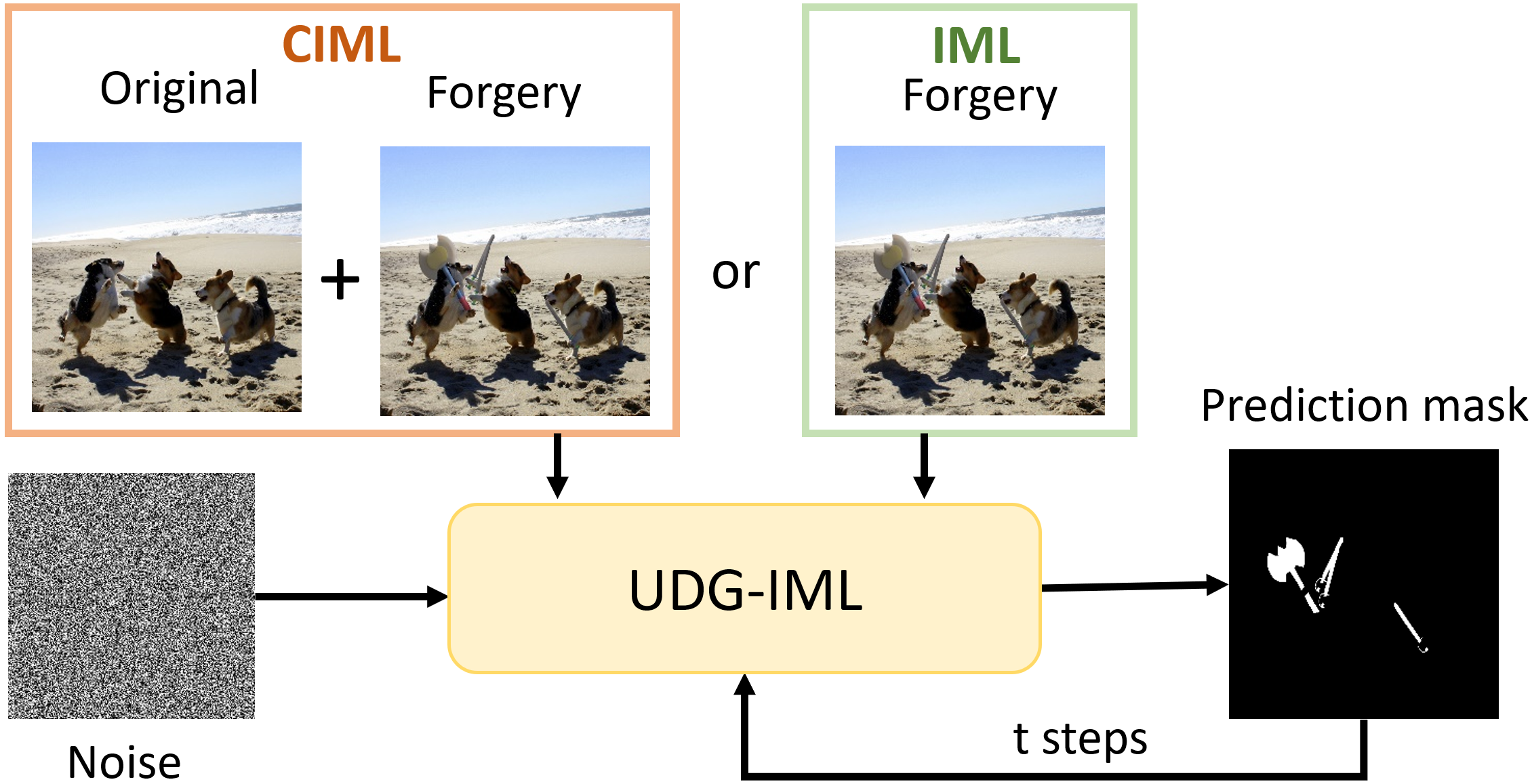}
	\caption{An overview of the proposed method. We can control whether the input image is a forgery image or a forgery image and a original image to control whether UGD-IML performs IML or CIML, thereby unifying the two tasks.}
	\label{fig:overview}
\end{figure}

In recent years, deep learning-based Image Manipulation Localization (IML) methods \cite{dong2022mvss,liu2022pscc,kwon2021cat,guo2023hierarchical} have shown great potential for identifying image manipulation regions.
These methods all rely heavily on high-quality manually labeled IML datasets \cite{dong2013casia,kniaz2019point,chen2021image,hsu2006detecting,guan2019mfc} to achieve high performance. However, limited by the size of the current datasets, these models perform poorly in generalizing to unseen data.
Therefore, a possible solution is to develop a large-scale generalized and comprehensive IML dataset covering various operating techniques in real-world scenarios. However, it is unrealistic to create such a dataset by manual annotation. 
To address this problem, several studies \cite{wu2017deep,liu2019adversarial,qu2024towards} have proposed the Constrained Image Manipulation Localization (CIML) methods to automatically obtain mask annotations of forged images from original and forged image pairs.
Early approaches \cite{wu2017deep,liu2019adversarial} employ a single association-based model to process all input data. However, due to the poor modeling, the obtained mask annotations of the forged images are often unreliable.
Although the recently proposed CAAA \cite{qu2024towards} is more effective, its structure is complex and the annotation process is cumbersome as it requires three different networks to be trained in advance to be used at different stages of the annotation. This complexity and the use of different models for related tasks is not in line with the future direction of the field. In addition, IML and CIML are essentially the same, just with different inputs, hence a unified and streamlined reusable model that can perform both CIML and IML tasks may be a good solution.

Current IML models \cite{dong2022mvss,liu2022pscc,kwon2021cat,guo2023hierarchical} are mainly based on discriminative models, which learn direct mapping relationships between input data and labels. Although discriminative-based models are simple and efficient, significant limitations lie in the heavy reliance on large amounts of manually labeled data, which limits the generalization ability of the models. In addition, discriminative methods lack uncertainty estimates to provide a confident assessment of the predicted results, and underperform especially in high-risk scenarios \cite{ji2023ddp}.
In contrast, the generative diffusion model reduces the need for labeled data by learning the latent distribution of the data and is able to provide uncertainty estimations to enhance the robustness and reliability of the predictions. Its stepwise denoising generative process leads to smoother and more consistent predictions, overcoming the shortcomings of discriminative methods.
Recently, generative diffusion models \cite{ho2020denoising,song2020denoising} have shown excellent performance on a variety of intensive prediction tasks, such as semantic segmentation \cite{li2021semantic,ji2023ddp} and depth estimation \cite{saxena2023monocular}. And since IML is also a class of pixel-level intensive prediction tasks, we believe that generative diffusion models can also provide an efficient, robust and reliable set of solutions for IML.

Based on the above analysis, we propose a novel Unified Generative Diffusion-based framework for IML and CIML tasks (UGD-IML), as shown in Fig. \ref{fig:overview}. 
Specifically, we map the ground truth to a high-dimensional continuous space by means of class embedding, and add Gaussian noise to this high-dimensional space to obtain the noise embeddings. 
Then we use an image encoder to obtain multi-scale features from input image and use FPN \cite{lin2017feature} to process these features to obtain the guidance conditions.
Finally the noise embeddings is combined with the guidance conditions to perform the diffusion process.
Moreover, our model can realize the conversion between IML (forgery image) and CIML (forgery image and original image) tasks by controlling the input image. Since we use a parameter-sharing module to encode the original and forgery images, our architecture can switch between IML and CIML tasks without introducing additional parameters and training. This innovation represents a major step forward in the field, providing a more efficient, robust, and reliable solution for image manipulation detection.

Our contributions can be summarized as follows:
\begin{itemize}
	\item We propose UGD-IML, the first unified framework for IML and CIML, enabling seamless execution of both tasks by simply controlling the input, without any architectural modifications.
	\item UGD-IML is a simple and efficient approach based on generative diffusion modeling. We reduce the computational complexity by decoupling the encoder from the decoder, so that the sampling process is performed only in the low-complexity decoder.
	\item We validate that our model achieves state-of-the-art performance on both IML and CIML through extensive experiments on multiple datasets.
\end{itemize}

\section{Related Works}
\subsection{Diffusion Models}
Diffusion model is a generative model that transforms noise into real data through a step-by-step denoising process, enabling efficient learning of the underlying distribution of the data.
DDPM \cite{ho2020denoising} proposes a diffusion model that add noise addition using a forward Markov process and denoise using a reverse Markov process. DDIM \cite{song2020denoising} presents non-Markovian sequences to speed up the sampling. More recently, numerous studies have expanded the application of diffusion models to broader tasks. DiffusionInst \cite{gu2024diffusioninst} and DiffusionDet \cite{chen2023diffusiondet} utilize the diffusion process for denoising object boxes and complete obeject detection. DDP \cite{ji2023ddp} applies the diffusion model framework to various dense tasks, such as depth estimation and semantic segmentation, allowing for the reuse of features over multiple steps. \cite{wang2023segrefiner} uses masks and RGB images as inputs to a diffusion model to perform discrete diffusion between the coarse mask and the ground truth, achieving the refinement of arbitrary segmentation masks. \cite{nam2023diffusion} effectively employs the diffusion model for dense matching between source and target images. 

\subsection{Image Manipulation Localization}
Most of the existing methods are based on handcrafted or learned fingerprint features to amplify the difference in contrast between the manipulated area and the normal area.
The current fingerprint feature-based methods are divided into two categories: noisy fingerprint-based methods and frequency-domain fingerprint-based methods.
Noise fingerprint-based methods \cite{cozzolino2019noiseprint,zhou2018learning,guillaro2023trufor} employ forgery fingerprints obtained from the noise level as an additional input to assist in localizing the manipulation regions.
In contrast, frequency-domain fingerprint-based methods \cite{kwon2021cat,li2019localization,qu2023towards} employ JPEG compression artifacts or anomalies in capturing high-frequency information as additional cues to localize the manipulation regions.
The introduction of additional fingerprint features makes these methods highly interpretable and perform well. However, they tend to show poor robustness on data outside the training distribution due to the constraints of fingerprint modal information.
To address this problem, some methods \cite{wu2019mantra,sun2023safl,chen2022learning} that utilize self-supervision, comparative learning, or simulated regression use confidence probabilities to determine high-risk manipulation regions.
This addresses the risk of overfitting posed by fingerprinting to some extent, but they come with the consequence of lower performance and higher training skill.
Recently, some work has attempted to explore the effectiveness of generative modeling in the IML domain.
For example, DiffForensics \cite{yu2024diffforensics} introduces a diffusion model which forces the model to learn intermediate features through pre-training and fine-tuning, and introduces edge losses to aid in model training. However, the capabilities of diffusion models are greatly limited due to the constraints of complex pre-training and multi-tasking structures.
DH-GAN \cite{LIU2024110658} use a novel GAN-based framework termed dual homology-aware generative adversarial network for image manipulation localization.
Although DH-GAN is based on GAN, the model employs the generator as a feature extractor in the latent space, which makes it logically fall under the category of discriminative models.
Therefore, the effectiveness of generative models in addressing localization tasks remains underexplored.

\begin{figure*}[ht] 
	\centering
	\includegraphics[width=16cm]{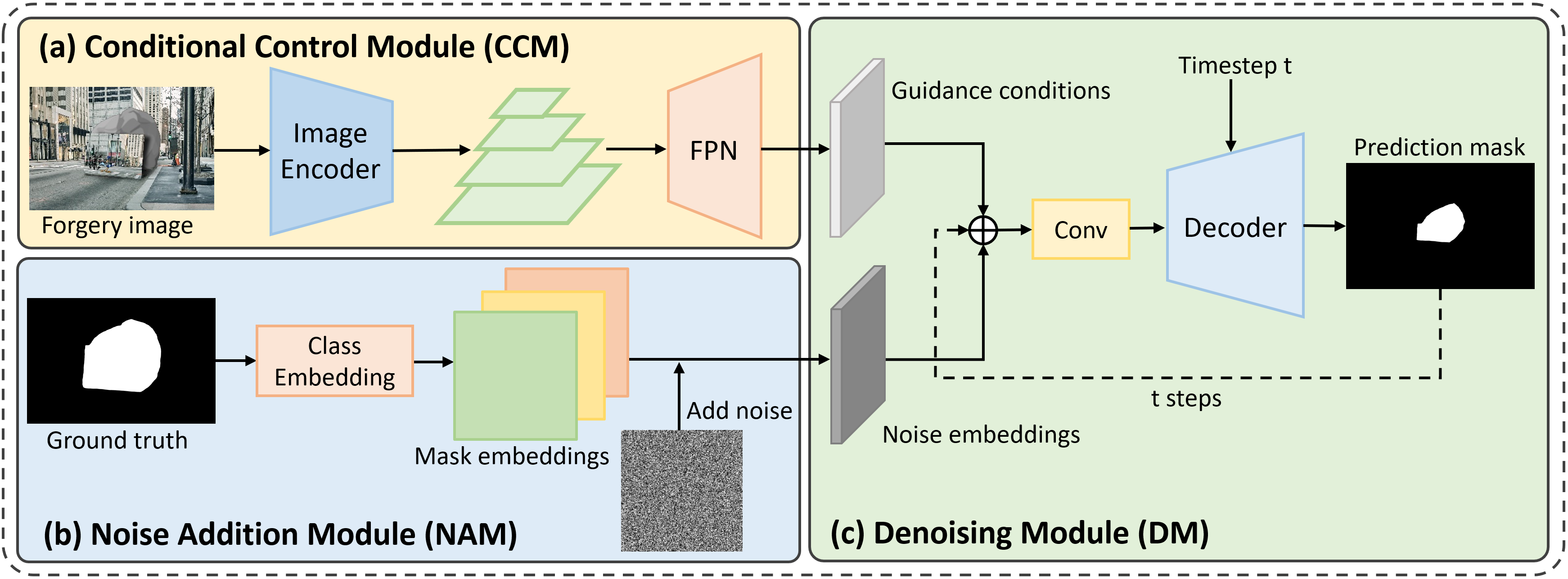} 
	\caption{The overall framework of UGD-IML. (a) The Conditional Control Module (CCM) performs feature extraction on the input image to get the guidance condutions.
    (b) The Noise Addition Module (NAM) performs class embedding on the ground truth and adds noise to get the noise embeddings. 
    (c) The Denoising Module (DM) takes guidance conditions and noise embeddings as input and generates the prediction mask guided by the guidance conditions.
	For IML, the forgery image $I^\text{Forg}$ is processed through the CCM and concatenated with the noise embeddings $x^\text{Mask}_t$. For CIML, the original image $I^\text{Org}$ is entered into the CCM with shared parameters as an additional condition, resulting in guidance conditions $c^\text{Org}$, which is then combined with guidance conditions $c^\text{Forg}$ and the noise embeddings $x^\text{Mask}_t$ for prediction, see Fig. \ref{fig:concatdetail} for more details.}
	\label{fig:overallframe}
\end{figure*}

\subsection{Constrained Image Manipulation Localization}
Unlike image processing localization, constrained image processing localization (CIML) requires reference to the original image to localize the forged image.
The concept of CIML is proposed by DMVN \cite{wu2017deep}, which detects similar objects in forging by sharing model parameters and correlation feature maps.
\cite{liu2019adversarial} proposes atrous convolution and adversarial training to locate splicing forging.
\cite{liu2020constrained} improves the performance of the model by introducing multiple attention modules to extract the attention-aware features.
Since the previous methods take a single correlation model for all input data , the performance of these methods is limited.
Recently, CAAA\cite{qu2024towards} classifies the forged and original image pairs into two categories, Shared Donor Group (SDG) and Shared Probe Group (SPG), by training a classifier based on whether the common part of the forgery image is a forged region or a real region. Then, a difference-aware semantic segmentation model and a semantic aligned correlation matching model are used to predict the manipulation regions in SPG and SDG, respectively.
Although CAAA has achieved better results in CIML, this staged, multi-model, automatic labeling method is cumbersome and very time consuming.


\section{Methodology}

\subsection{Preliminaries}
\label{sec:3.1}
Conditional diffusion models integrate condition $c$ (e.g., class labels, text embeddings, or structured data), enabling vanilla diffusion models~\cite{ho2020denoising} to explicitly control over the generative trajectory during both training and sampling phases. Normally, a conditional diffusion model involves two processes: a forward process that destroys the image into noise, and a reverse process that restores the noise to the image.

Given a timestep $t \sim \text{Uniform}(1, T)$, noise schedule $\beta_t \in (0, 1)$ can directly add the Gaussian noise from $x_0 \sim p(x_0)$ to $x_t$:
\begin{equation}
	x_t = \sqrt{\bar{\alpha}_t} x_0 + \sqrt{1 - \bar{\alpha}_t}\epsilon_t,
	\label{eq:add}
\end{equation}
where $\bar{\alpha}_t = \prod_{i=1}^{t} \alpha_i = \prod_{i=1}^{t} (1 - \beta_i)$, and $\epsilon_t \sim \mathcal{N}(0, I)$ is a Gaussian noise. The diffusion model $f_{\theta}(x_t, c, t)$ estimates $\epsilon_t$ under the guidance of condition $c$ and timestep $t$ by minimizing the training loss:
\begin{equation}
	\mathcal{L}_\text{diffusion} = \mathbb{E}_{x_0, \epsilon_t, t}\| \epsilon_t - f_{\theta}(x_t, c, t) \|_2^2.
	\label{eq:loss1}
\end{equation}

If $f_\theta(x_t, c, t)$ has been trained, Eq.~\ref{eq:sampling1} can be iterated repeatedly to obtain $x_0$ from a Gaussian noise $x_T \sim \mathcal{N}(0,I)$:
\begin{equation}
	x_{t-1} = \frac{1}{\sqrt{\alpha_t}}(x_t - \frac{1 - \alpha_t}{\sqrt{ 1 -\bar{\alpha_t}}}f_\theta(x_t, c, t)) + \sigma_tz,
	\label{eq:sampling1}
\end{equation}
where $\sigma_t$ is usually set $\beta_t$ or $\frac{1 - \alpha_t}{\sqrt{ 1 -\bar{\alpha_t}}}\beta_t$, and $z \sim \mathcal{N}(0,I)$ is also a Gaussian noise.

\subsection{Overall Architecture of UGD-IML}
UGD-IML fully follows the structure of conditional diffusion modeling, which guides the generation of the model by using input image as conditions, as shown in Fig.~\ref{fig:overallframe}. 
In particular, UGD-IML consists of three modules: Condition Control Module (CCM), Noise Addition Module (NAM) and Denoising Module (DM).

\begin{figure}[t]
	\centering
	\includegraphics[width=7cm]{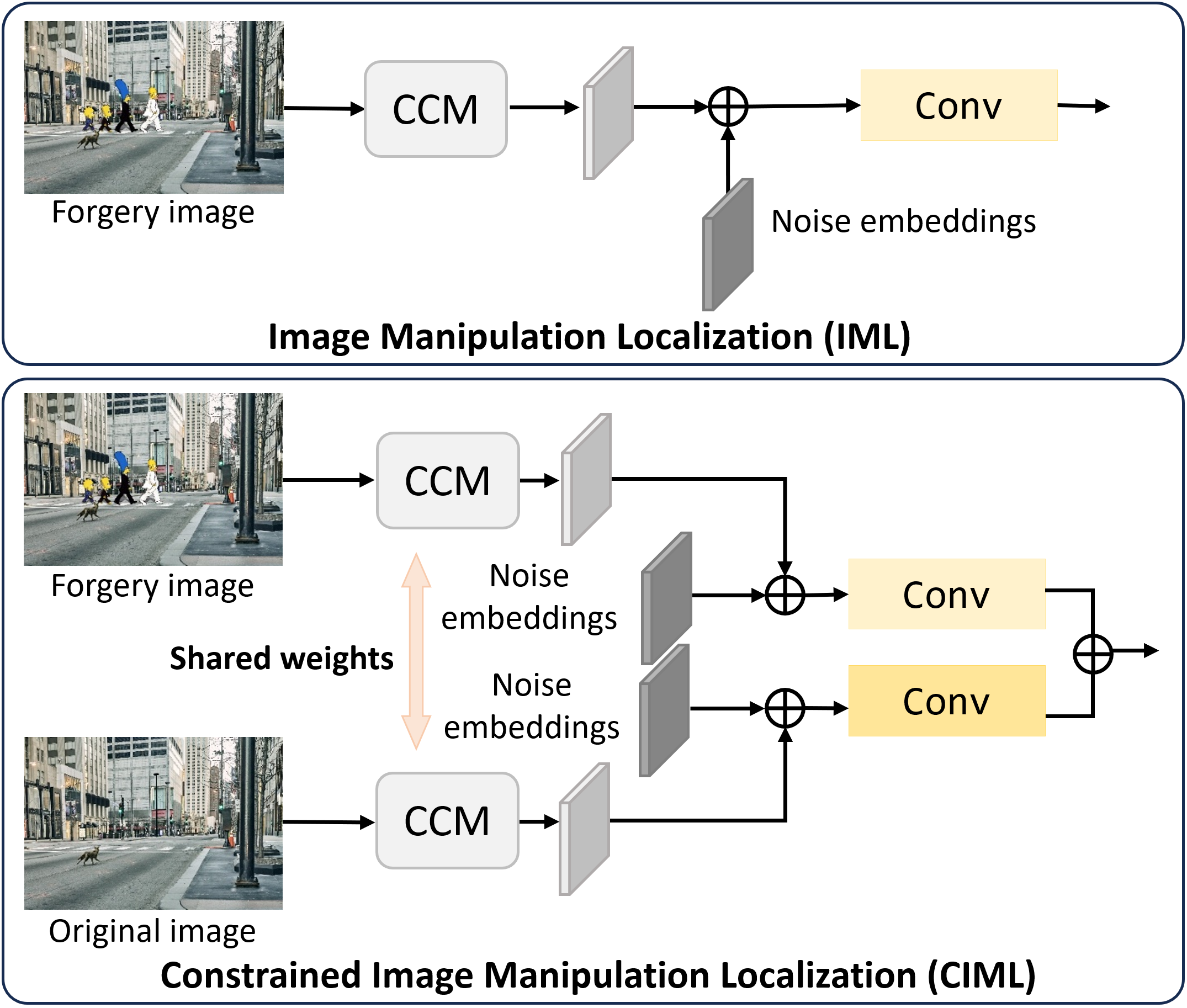}
	\caption{Illustration of the CCM's different processes for IML and CIML tasks.}
	\label{fig:concatdetail}
\end{figure}

\noindent\textbf{Input Image Guidance.} 
We introduce a Conditional Control Module (CCM) that determines whether the model performs IML or CIML by controlling the input image, thereby guiding the generation trajectory of the diffusion model, as shown in Fig.~\ref{fig:concatdetail}.
UGD-IML performs IML when the input is only a forgery image $I^\text{Forg}$. Specifically, we employ the Swin-Transformer-based \cite{liu2021swin} image encoder to extract multi-scale features of the image at different resolutions. 
These multi-scale features are then processed via a Feature Pyramid Network (FPN) \cite{lin2017feature} to obtain guidance conditions $c^\text{Forg}$. CCM can be described as:
\begin{equation}
	c^\text{Forg} = \text{FPN}(\text{img\_encoder}(I^\text{Forg})),
\end{equation}
where $\text{img\_encoder}(\cdot)$ is the image encoder.

In contrast, UGD-IML performs CIML when the inputs are forgery images $I^\text{Forg}$ and original images $I^\text{Org}$.
Similar to the processing flow of IML, CIML also requires that the input image is first encoded using image encoder and then processed using FPN to obtain guidance conditions $c^\text{Forg}$ and $c^\text{Org}$. 
The difference is that CIML employs CCM with shared weights to extract the feature maps of the two images separately, and then splices them with the noise maps output from NAM respectively.
By using the same modeling framework in both IML and CIML, we can reuse models, which not only greatly reduces the number of model references, but also makes the whole process much simpler and easier.

\begin{algorithm}[htbp]
    \caption{Training a UGD-IML for IML} 
    \label{alg:training}
    \small
    \begin{algorithmic}[1] 
        \Repeat
        \State $(I^\text{Forg}, gt^\text{Mask}) \sim p(I^\text{Forg}, gt^\text{Mask})$
        \State $c^\text{Forg} = \text{FPN}(\text{img\_encoder}(I^\text{Forg}))$
        \State $x^\text{Mask}_0 = \text{Norm}(\text{class\_embedd}(gt^\text{Mask}))$
        \State $t \sim \text{Uniform}(1,T)$
        \State $\epsilon_t \sim \mathcal{N}(0, I)$
        \State $x^\text{Mask}_t = \sqrt{\bar{\alpha}_t} x^\text{Mask}_0 + \sqrt{1 - \bar{\alpha}_t}\epsilon_t$
        \State $\hat{x}^\text{Mask}_0 = \text{DM}(x^\text{Mask}_0, c^\text{Forg}, t)$
        \State Take gradient descent step on 
        \Statex $\qquad \qquad \mathcal{L} = \lambda \mathcal{L}_{\text{wce}} + (1 - \lambda) \mathcal{L}_{\text{dice}}$
        \Until converged
    \end{algorithmic}
\end{algorithm}

\noindent\textbf{Training Process.}
The mathematical structure of the diffusion model inherently supports continuous data, which makes it particularly well suited to the regression task of predicting continuous-type values. 
However, IML and CIML are discrete tasks, and several studies \cite{ji2023ddp,chen2023generalist} have shown that standard diffusion models perform poorly on discrete tasks.
To make the diffusion model suitable for our task, we use class embedding in Noise Addition Module (NAM) to transform discrete ground truth $gt^\text{Mask}$ into a high-dimensional continuous space and normalize them to obtain mask embeddings $x_0^\text{Mask}$, as follows:
\begin{equation}
    x^\text{Mask}_0 = \text{Norm}(\text{class\_embedd}(gt^\text{Mask})).
\end{equation}
As previously mentioned, we then add Gaussian noise $\epsilon_t$ to $x_0^\text{Mask}$ by Eq.~\ref{eq:add} to obtain noise embeddings $x_t^\text{Mask}$.

In denoising module (DM), $x_t^\text{Mask}$ and $c^\text{Forg}$ (and $c^\text{Org}$) still require further processing before being input to the decoder. 
For IML, $x_t^\text{Mask}$ and $c^\text{Forg}$ are connected along the channel axis and then fused by a convolution layer. In contrast, $c^\text{Forg}$ and $c^\text{Org}$ are respectively processed in the same way as IML and then concatenated together for CIML, as shown in Fig~\ref{fig:concatdetail}.
We choose the DETR \cite{lin2024dcea} model based on deformable attention as the decoder, because the deformable-attention-based decoder architecture shows excellent performance as well as low computational complexity in a variety of image localization tasks \cite{ji2023ddp,cheng2022masked,zhang2022dino}.
And also unlike the vanilla diffusion model~\cite{ho2020denoising}, the decoder outputs the predicted mask $\hat{x}^\text{Mask}_0$ instead of noise $\epsilon_t$.
Because studies~\cite{ye2025bevdiffuser, ji2023ddp} have indicates that compared with the loss function defined in Eq.~\ref{eq:loss1}, using task-specific loss is more effective for supervision. 
Therefore, we use the loss function commonly used in image manipulation localization. Specifically, the loss function is a linear combination of the weighted cross-entropy $\mathcal{L}_{\text{wce}}$ and the dice loss $\mathcal{L}_{\text{dice}}$, i.e.:
\begin{equation}
	\mathcal{L}_\text{dice} = 1 - \frac{2 \times |gt^\text{Mask} \cap \hat{x}^\text{Mask}_0|}{|gt^\text{Mask}| + |\hat{x}^\text{Mask}_0|}
	,
\end{equation}
\begin{equation}
	\mathcal{L}_\text{wce} = \mu gt^\text{Mask} \log(\hat{x}^\text{Mask}_0) + \eta (1 - gt^\text{Mask}) \log(1 - \hat{x}^\text{Mask}_0),
\end{equation}
\begin{equation}
	\mathcal{L} = \lambda \mathcal{L}_{\text{wce}} + (1 - \lambda) \mathcal{L}_{\text{dice}},
\end{equation}
\begin{table*}[hbtp]
	\centering
	\setlength{\aboverulesep}{0pt}
	\setlength{\belowrulesep}{0pt}
	\renewcommand\arraystretch{1}
	\setlength{\tabcolsep}{5.5pt}
	\small
	\begin{tabular}{l|c|cccccc|cc|c} 
		\toprule
		\multirow{2}{*}{\textbf{Methods}} & \multirow{2}{*}{\textbf{Source}} & \multicolumn{6}{c|}{\textbf{Editing}} & \multicolumn{2}{c|}{\textbf{DGM}} & \multirow{2}{*}{\textbf{ \itshape Average}} \\
		&  & CASIA1.0+ & Columbia & NIST16 & IMD2020 & DSO-1 & Korus & AutoSpl. & OpenFor. & \\
		\hline
		H-LSTM & \scriptsize \itshape TIP, 2019 & 0.121 & 0.257 & 0.109 & 0.118  & 0.187 &0.089  & 0.306  & 0.123  & 0.164 \\
		ManTra-Net  & \scriptsize \itshape CVPR, 2019 & 0.136 & 0.357 & 0.160 & 0.180 & 0.089 & 0.104 & 0.192 & 0.043 & 0.158\\
		SPAN & \scriptsize \itshape ECCV, 2020  & 0.088 & 0.213 & 0.116 & 0.108 & 0.059 & 0.070 & 0.047 & 0.014 & 0.089\\
		CAT-Net & \scriptsize \itshape WACV, 2021  & 0.394 & 0.854 & 0.336 & 0.295 & 0.135 & 0.149 & 0.185 & 0.003 &0.294\\
		SATL-Net & \scriptsize \itshape TIFS, 2022  & 0.064 & 0.677 & 0.175 & 0.142 & 0.084 & 0.039 & 0.103 & 0.019 &0.163\\
		MVSS-Net & \scriptsize \itshape TPAMI, 2022  & 0.451 & 0.665 & 0.292 & 0.264 & 0.271 & 0.095 & 0.333 & 0.056 &0.303\\
		PSCC-Net & \scriptsize \itshape TCSVT, 2022  & 0.355 & 0.672 & 0.238 & 0.295 & 0.318 & 0.156 & 0.150 & 0.065 &0.281\\
		HiFi-Net & \scriptsize \itshape CVPR, 2023 & 0.092 & 0.382 & 0.172 & 0.178 & 0.304 & 0.088 & \textbf{0.613} &\textbf{0.149} & 0.247\\
		DiffForensics & \scriptsize \itshape CVPR, 2024 & \underline{0.517} & \underline{0.912} & \underline{0.415} & \underline{0.511} & \underline{0.485} & \underline{0.257} & 0.507 &0.122 & \underline{0.466}\\
		\hline
		\textbf{UGD-IML} & \itshape \textbf{ours}  & \textbf{0.603} & \textbf{0.949} & \textbf{0.451} & \textbf{0.588} & \textbf{0.580} & \textbf{0.290} &  \underline{0.573} &  \underline{0.134} & \textbf{0.511}\\
		\hdashline
		\multicolumn{2}{c|}{\itshape improvement to existing best}  & \scriptsize $16.63\%$ & \scriptsize $4.06\%$ & \scriptsize $8.67\%$ & \scriptsize $15.07\%$ & \scriptsize $19.59\%$ &\scriptsize $ 12.84\%$ &  - &  - & \scriptsize $9.66\%$  \\
		\bottomrule
	\end{tabular}
	\caption{Pixel-level F1 results for Image Manipulation Localization (IML). Best in bolded and second in underlined.}
	\label{tab:IMLperformance}
\end{table*}
where $\cap$ calculates the intersection between $x^\text{Mask}_0$ and $gt^\text{Mask}$, $|\cdot|$ computes the sum of the elements, the class weight $\mu$ is set to 0.5 and $\eta$ is set to 2.5 for pixel class balance, and the loss weight $\lambda$ is set to 0.3. The complete training process is shown in Algorithm \ref{alg:training}.

\noindent\textbf{Sampling Process.}
Under the guidance of condition obtained by processing the input image via CCM, DM is iterated to obtain $x^\text{Mask}_0$ starting from a Gaussian noise $x^\text{Mask}_T$, as shown Algorithm \ref{alg:sampling}. 
The low-complexity decoder greatly reduces the computational burden, so it allows the sampling to proceed efficiently.
Different from Eq.~\ref{eq:sampling1}, we use the DDIM sampler~\cite{song2020denoising}.
It enables DM to sample along deterministic generative trajectories, further accelerating the process without stochasticity. Given a timestep $\tau_s \in [\tau_1, \tau_2, ..., \tau_S]$ that is a sub-sequence of $[1, 2, ..., T]$ in which $S \ll T$, the DDIM sampler can be described as:
\begin{equation}
	\hat{x}^\text{Mask}_0 = \text{DM}(x^\text{Mask}_{\tau_s}, c, \tau_s),
\end{equation}
\begin{equation}
	x_{\tau_{s-1}} = \sqrt{\Bar{\alpha}_{\tau_{s-1}}}\hat{x}^\text{Mask}_0 + \frac{\sqrt{1- \Bar{\alpha}_{\tau_{s-1}}}}{\sqrt{1- \Bar{\alpha}_{\tau_s}}}(x_{\tau_s} - \sqrt{\Bar{\alpha}_{\tau_s}}\hat{x}^\text{Mask}_0).
\end{equation}
where $c$ is $c^\text{Forg}$ for IML, or is $c^\text{Forg}$ and $c^\text{Org}$ for CIML.

\begin{algorithm}[htbp]
    \caption{Sampling form a UGD-IML for IML}
    \label{alg:sampling}
    \small
    \begin{algorithmic}[1] 
    \Require Forgery image $I^\text{Forg}$
    \State $c^\text{Forg} = \text{FPN}(\text{img\_encoder}(I^\text{Forg}))$
    \State $x^\text{Mask}_T \sim \mathcal{N}(0,I)$
    \For{$\tau_S, \tau_{S-1}...,\tau_1$}
    \State $\hat{x}^\text{Mask}_0 = \text{DM}(x^\text{Mask}_{\tau_s}, c^\text{Frog}, \tau_s)$
    \State $x_{\tau_{s-1}} = \sqrt{\Bar{\alpha}_{\tau_{s-1}}}\hat{x}^\text{Mask}_0 + \frac{\sqrt{1- \Bar{\alpha}_{\tau_{s-1}}}}{\sqrt{1- \Bar{\alpha}_{\tau_s}}}(x_{\tau_s} - \sqrt{\Bar{\alpha}_{\tau_s}}\hat{x}^\text{Mask}_0)$
    \EndFor
    \State \textbf{return} {$x^\text{Mask}_0$}
    \end{algorithmic}
\end{algorithm}

\section{Experiments}

\subsection{Image Manipulation Localization}
\subsubsection{Datasets.}
During training, we use the CASIAv2 \cite{dong2013casia} with 7491 real and 5105 fake images and the Fantasitic-Realiy \cite{kniaz2019point}  with 16592 real and 19423 fake images to train UGD-IML. 
\begin{table}[h]
	\centering
	\small
	\setlength{\aboverulesep}{0pt}
	\setlength{\belowrulesep}{0pt}
	\renewcommand\arraystretch{1}
	\setlength{\tabcolsep}{6.6pt}
	\begin{tabular}{ccccc}
		\toprule
		Steps & CASIA & Columbia & DSO-1 & \itshape Average\\
		\hline
		0   & 0.551 & 0.920 & 0.563  & 0.678\\
		1   & 0.603 & 0.949 & 0.580 & 0.714\\
		3   & 0.605 & 0.949 & 0.581 & 0.715\\
		\bottomrule
	\end{tabular}
	\caption{Ablation results on Diffusion.} 
	\label{tab:imldiff}
\end{table}
As for the test datasets, we use various challenging datasets to verify the generalization and capabilities of the model, including CASIAv1+ \cite{chen2021image}, Columbia \cite{hsu2006detecting}, DSO-1 \cite{de2013exposing}, IMD2020 \cite{novozamsky2020imd2020}, NIST16 \cite{guan2019mfc} and Korus \cite{korus2016multi}. 
We further test on datasets forgery by Deep Generative Model (DGM), including AutoSplicing \cite{jia2023autosplice} and OpenForensics \cite{le2021openforensics}.

\subsubsection{Evaluation Metrics.}
Since different thresholds provide inconsistent comparisons, to provide a fair comparison with the baseline methods \cite{yu2024diffforensics}, we choose a measure consistent with it, namely F1 with a threshold of 0.5.

\subsubsection{Baseline Methods.}
To ensure fairness and accuracy of the comparison, we selected only state-of-the-art methods for which the authors provided pre-trained models, released the source code, or evaluated them against common standards.
Specifically, we compare the proposed method with several current state-of-the-art methods: MantraNet \cite{wu2019mantra}, SPAN \cite{hu2020span}, MVSS-Net \cite{dong2022mvss}, HiFi-Net \cite{guo2023hierarchical}, PSCC-Net \cite{liu2022pscc}, CAT-Net \cite{kwon2021cat} and DiffForensics \cite{yu2024diffforensics}.

\begin{figure*}[ht] 
	\centering
	\includegraphics[width=17cm]{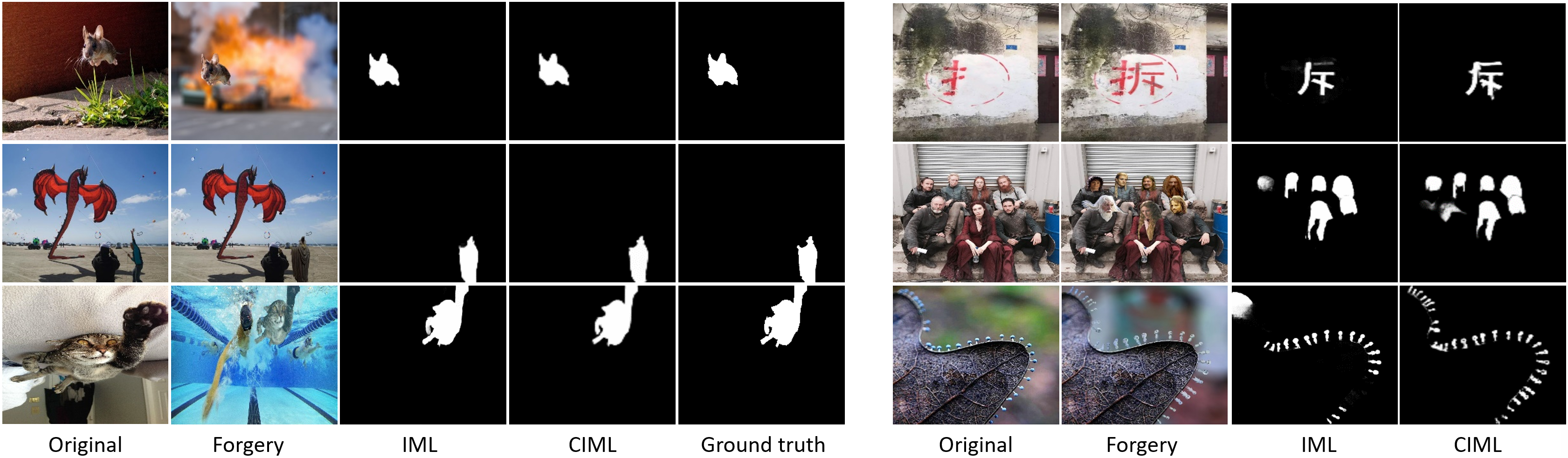}
	\caption{The visualization results of our method are in IML and CIML. The images on the left are from CASIA and IMD testing sets. The images on the right are from several challenging wild samples available on the Internet.}
	\label{fig:qulity reults}
\end{figure*}

\subsubsection{Implementation Details.}
The network adopts Swin-B \cite{liu2021swin} as the encoder and a six-layer deformable DETR decoder \cite{lin2024dcea}, with learnable ground-truth class embeddings.
In data preprocessing, randomly resizing and cropping the input image to $512\times 512$ can greatly improve the model's detection ability, but for a fair comparison with previous methods, we use the same settings as in DiffForensics \cite{yu2024diffforensics}, i.e., resizing the input image to $512\times 512$ and randomly applying jpeg compression from 30 to 100.
In training, we set the training step size $T$ to 1000 and use DDIM for the inverse process. 
To ensure fairness and speed, all our subsequent experiments and results are realized with DDIM step size 1 without skiiping step. In addition, the batch size is set to 12, the learning rate is 6e-5, and 50 epochs are trained using AdamW as the optimizer.

%
%

\begin{table}[t]

	\centering
	\small
	\setlength{\aboverulesep}{0pt}
	\setlength{\belowrulesep}{0pt}
	\renewcommand\arraystretch{1}
	\setlength{\tabcolsep}{5pt}
	\begin{tabular}{c|cc|cc}
		\toprule
		\multirow{2}{*}{Methods} & \multicolumn{2}{c|}{IMDSDG} & \multicolumn{2}{c}{IMDSPG} \\
		\cline{2-5}
		
		& F1 & IOU & F1 & IOU \\
		\hline
		DMVN*   & 0.434 & 0.276 & 0.430 & 0.276 \\
		DMAC*  & 0.559 & 0.410 & 0.620 & 0.432  \\
		\textbf{UGD-IML*}  & \underline{0.811} & \underline{0.727} & 0.801 & 0.700\\
		\hline
		DMVN   & 0.495 & 0.317 & 0.728 & 0.578 \\
		DMAC  & 0.660 & 0.518 & 0.728 & 0.537  \\
		CAAA  & 0.798 & 0.702 & \underline{0.889} & \underline{0.834} \\
		\textbf{UGD-IML}  & \textbf{0.854} & \textbf{0.780} & \textbf{0.904} & \textbf{0.855}\\
		\hdashline
		\itshape improvement to existing best  & \scriptsize $7.02\%$ & \scriptsize $11.11\%$ & \scriptsize $1.69\%$&\scriptsize $2.52\%$  \\
		\bottomrule
	\end{tabular}	
	\caption{Pixel-level F1 and IOU results for Constrained Image Manipulation Localization (CIML).  * denotes training using both SDG and SPG data.}
	\label{tab:cimlperformence}
\end{table}
\subsubsection{Comparison Experiments.}
As shown in Tab. \ref{tab:IMLperformance}, our method achieves excellent performance on all datasets.
Compared to the current best method, DiffForensics \cite{yu2024diffforensics}, our method achieves a significant performance improvement on all datasets with the same parameter scales (average improvement of $9.66\%$).
Moreover, our method achieves state-of-the-art performance compared to all baseline models on all datasets with editing forging (average improvement of  $12.81\% $).
On the DGM datasets, our method still obtains a significant improvement compared to DiffForensics, but HiFi-Net \cite{guo2023hierarchical} achieves the best performance. This is due to the fact that HiFi-Net is specifically designed for generative forging, so it has a poor performance on edit-forgery datasets.
Specifically, on the editing dataset, our model improves on average compared to HIFI by $236\% $.
Taken together, our model has the best performance compared to all current models.

\subsubsection{Ablation on Diffusion.}
Our method is a typical conditional diffusion model with a forward noise addition and a reverse denoising process.
In order to demonstrate the effectiveness of the diffusion process in our task, we perform ablation studies.
We first compare the effect of UGD-IML with the conventional decoder-encoder structure without the diffusion process. 
\begin{table}[htbp]
	\centering
	\small
	\setlength{\aboverulesep}{0pt}
	\setlength{\belowrulesep}{0pt}
	\renewcommand\arraystretch{1}
	\begin{tabular}{ccccc}
		\toprule
		\multirow{2}{*}{Steps} & \multicolumn{2}{c}{IMDSDG} & \multicolumn{2}{c}{IMDSPG} \\
		\cline{2-5}
		& F1 & IOU & F1 & IOU \\
		\hline
		0   & 0.793 & 0.691 & 0.752 &0.660  \\
		1   & 0.811 & 0.727 & 0.801 &0.700 \\
		3   & 0.813 & 0.728 & 0.802 &0.700  \\
		\bottomrule
	\end{tabular}
	\caption{Ablation results on diffusion.} 
	\label{tab:cimldiff}
\end{table}
Specifically, we set all the added noise to 0 and keep the network architecture unchanged.
In addition, since the effect of the model prediction is affected by the number of diffusions, we further validate the effect of different numbers of diffusions on the experimental results. 
The experimental results are shown in Tab. \ref{tab:imldiff}.
From the results, it can be seen that our proposed conditional diffusion-based UGD-IML has a significant improvement in effectiveness compared to the structure without diffusion process. In addition, increasing the number of diffusion steps does not yield an effective enhancement, so all the results in this paper are based on single-step diffusion.

\subsection{Constrained Image Manipulation Localization.}
\subsubsection{Datasets.}
We use CASIAv2 \cite{dong2013casia}, IMD2020 \cite{novozamsky2020imd2020} and COCO dataset \cite{lin2014microsoft} for training.
For IMD2020, we follow the strategy in CAAA \cite{qu2024towards}, i.e., we classify the forged images in IMD20 into two categories, SPG and SDG, and randomly divide them into training and testing sets using the classifier provided by CAAA in a 3:1 ratio.
Since IMD can be divided into relatively balanced SDG and SPG image pairs, while the other test sets are very unbalanced. Therefore our tests are also performed on IMDSDG and IMDSPG.

\subsubsection{Baseline Methods.}
We compare the three most representative of the constrained image localization methods: DMVN \cite{wu2017deep}, DMAC \cite{liu2019adversarial},  CAAA \cite{qu2024towards}. 

\begin{figure*}[htbp]
	\centering
	\includegraphics[width=17cm]{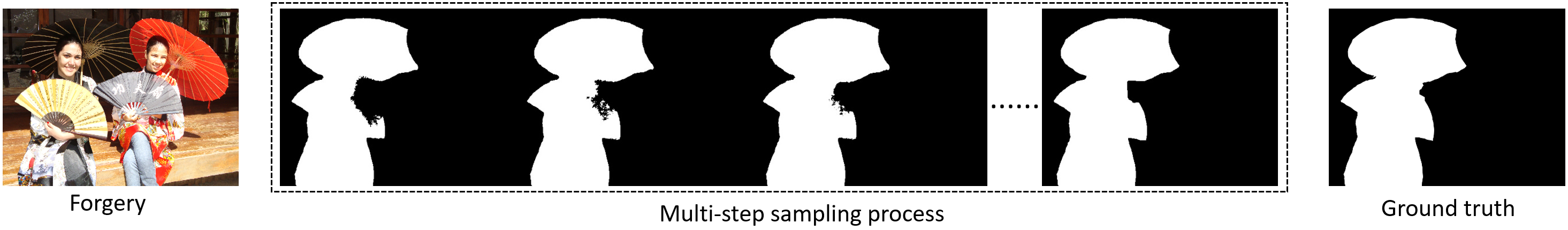} 
	\caption{Visualization of the results of a multi-step sampling process. For more visualization, we use larger sampling steps here. In practical reasoning, we use just one step or a few steps.}
	\label{fig:visualdiff}
\end{figure*}

\subsubsection{Evaluation Metrics.}
Pixel-level F1 and AUC (Area Under Curve of a ReceiverOperating-Characteristic curve) are used as evaluation Metrics.
As with IML, the threshold for F1 is still set to 0.5.
To account for randomness in IMD partitioning, we repeated the experiment three times with different random splits, retraining and testing each time. The final results are averaged over the three runs.

\subsubsection{Implementation Details.}
The CIML model structure largely follows that of IML. Unlike IML, which uses only the forgery image as input, CIML feeds both the forgery and original images into a shared-parameter CCM for encoding.
In addition, we set the batch size to 6 in training.
To better align with the previous method, we resize the input image to $512\times 512$. 
Notably, we also use an inverse step of 1.
The CIML setup and training strategy is similar to the IML task, which means that our model can be switched between the IML and CIML tasks at will without any modification.

\subsubsection{Comparison Experiments.}
As shown in Tab. \ref{tab:cimlperformence}. 
The performance of all the models becomes worse, when trained with both SDG and SPG data.
Since CAAA models use different networks for different types of data, CAAA does not support training with both SDG and SPG data.
When trained with both types of data simultaneously, our model demonstrates significant improvements compared to DMVN and DMAC. Moreover, it achieves notable performance gains even when compared to DMVN and DMAC trained separately on individual data types. Additionally, on the IMDSDG dataset, our method is better than the current best performing CAAA.
When training UGD-IML separately for different data, our method achieves state-of-the-art performance on both datasets.
Specifically, compared to CAAA, our method improves F1 and IOU by $7.02\%$, $11.11\%$ and $1.69\%$, $2.52\%$ on IMDSDG and IMDSPG, respectively.
It is worth noting that our approach completes the task in a single step, whereas CAAA requires multiple steps involving three separate models. In contrast to the multi-model, multi-step framework of CAAA, our model is both reusable and more parameter-efficient, making our method significantly more practical and efficient.

\subsubsection{Ablation on Diffusion.}
The experimental setup here is the same as in IML. As shown in Tab. \ref{tab:cimldiff}, it can be concluded that diffusion also plays a crucial role in CIML. And the performance of the model is only slightly improved after increasing the number of sampling steps.


\subsection{Visualization Results}

\begin{figure}[t]
	\centering
	\includegraphics[width=8cm]{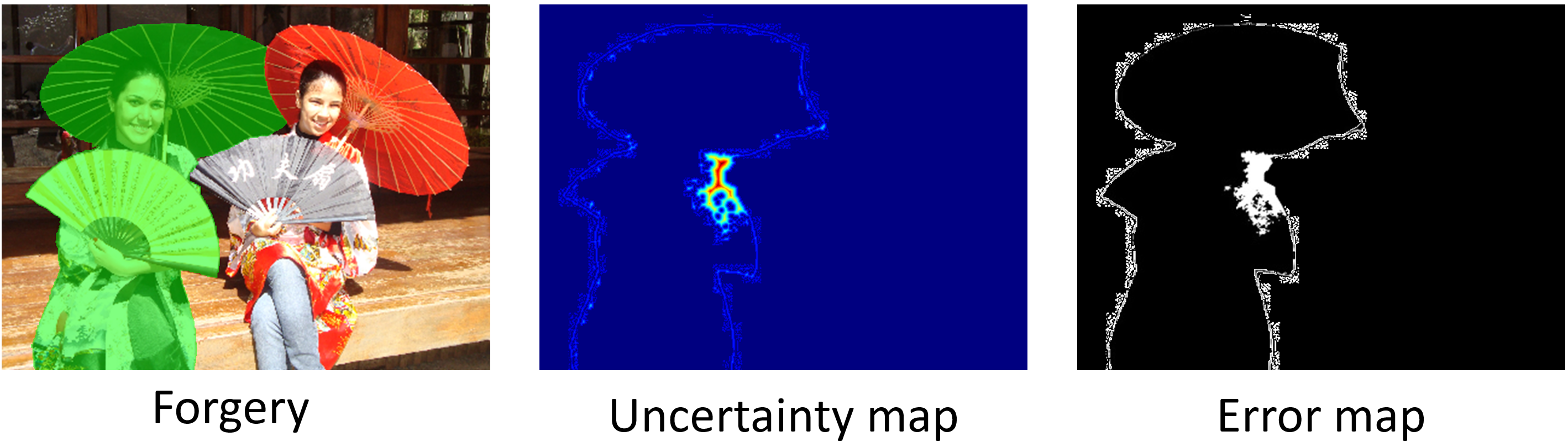}
	\caption{Visualization of uncertainty awareness. Bright regions in the uncertainty map reflect high uncertainty, while white areas in the error map indicate misclassified points.
}
	\label{fig:compress}
\end{figure}

\subsubsection{Qualitative Results.}
To visualize the localization effectiveness of our methods, we show the prediction masks of our methods on the IML and CIML tasks. As shown in Fig. \ref{fig:qulity reults}.
The visualization results show that our methods are not only able to accurately locate the forgery regions, but also output clear forgery boundaries for both IML and CIML.
Moreover, the good performance on challenging wild data indicates that our method has very good generalization.

\subsubsection{Dynamic Inference.}
The inference of the diffusion model is controlled by the sampling step size, and the results obtained by different step sizes are somewhat different. We can get more accurate localization effect by increasing the sampling step size. As shown in Fig. \ref{fig:visualdiff}. As the step size increases, the prediction of the model gets closer and closer to the ground truth.
This is also known as the dynamic inference property of the diffusion model, which means that our model can balance the computational cost and accuracy of the model by controlling the step size without the need of additional parameters and training.

\subsubsection{Uncertainty Awareness.}
During the multi-step sampling process, we count the pixels where the prediction result of each step is different from the result of the previous step. We then normalize this change count mapping to 0-1 to obtain an uncertainty mapping. As shown in Fig. \ref{fig:compress}.
This suggests that our method is able to easily estimate forecast uncertainty, a capability that previous IML models lack.

\section{Conclusion}
This paper introduces UGD-IML, an innovative framework that effectively integrates IML and CIML tasks. Extensive experiments demonstrate its superior performance on various datasets, highlighting the advantages of diffusion-based architectures in image manipulation localization. The model's efficiency, robustness, and uncertainty estimation capabilities enable it to address complex real-world scenarios effectively, providing reliable technical support for image authenticity detection. UGD-IML represents a significant step forward in this field, with promising potential for broader applications and continued optimization.


\bibliography{main}


\end{document}